\documentclass{article}
\usepackage{apacite}

\usepackage[preprint]{nips_2018}

\usepackage[utf8]{inputenc} 
\usepackage[T1]{fontenc}    
\usepackage{hyperref}       
\usepackage{url}            
\usepackage{booktabs}       
\usepackage{amsfonts}       
\usepackage{nicefrac}       
\usepackage{microtype}      
\usepackage{amsmath} 
\usepackage{amssymb}
\usepackage{graphicx}
\usepackage{subfigure}
\usepackage{geometry}
\usepackage{float}

\usepackage{mathrsfs}

\title{O-GAN: Extremely Concise Approach for Auto-Encoding Generative Adversarial Networks}

\author{
  Jianlin Su \\
  School of Mathematics\\
  Sun Yat-sen University\\
  Guangdong, China \\
  \texttt{bojone@spaces.ac.cn} \\
}

\begin{document}

\maketitle

\begin{abstract}
In this paper, we propose Orthogonal Generative Adversarial Networks (O-GANs). We decompose the network of discriminator orthogonally and add an extra loss into the objective of common GANs, which can enforce discriminator become an effective encoder. The same extra loss can be embedded into any kind of GANs and there is almost no increase in computation. Furthermore, we discuss the principle of our method, which is relative to the fully-exploiting of the remaining degrees of freedom of discriminator. As we know, our solution is the simplest approach to train a generative adversarial network with auto-encoding ability.
\end{abstract}

\section{Introduction}

Generative Adversarial Networks (GANs) have made great success on image synthesis since its first work \citep{Goodfellow2014Generative}. Nowadays we have serveral famous GANs, such as PGGAN \citep{Karras2017Progressive}, BigGAN \citep{Brock2018Large} and StyleGAN \citep{Karras2018A}, which can generate high resolution images in various scenarios. 

Typically, a GAN consists of two networks: generator and discriminator (aka critic). In other words, a typical GAN is lack of a inferring model (aka encoder). And we also have another popular kind of generative models, Variational Autoencoders (VAEs, \cite{Kingma2013Auto}), which can provide us both encoder and generator. However, images generated by VAEs are always lack of fidelity.

 A large class of GAN variants aim to combine ideas from VAEs and GANs, such as variational auto-encoder GANs (VAE-GANs, \cite{Larsen2015Autoencoding}), adversarial generator encoders (AGEs, \cite{Ulyanov2017Adversarial}) and auto-encoding GANs ($\alpha$-GANs, \cite{Rosca2017Variational}). Some other models, such as Adversarial Autoencoders (AAEs, \cite{makhzani2015adversarial}), Bidirectional GANs (BiGANs, \cite{Donahue2016Adversarial}) and Adversarially Learned Inference (ALI, \cite{Dumoulin2016Adversarially}), are adding a encoder into generative model under the framework of GAN directly. 

All of these GAN variants have at least three components: generator, discriminator and encoder. The discriminator and encoder have similar network architecture. However, as the the training procedure passes by, the generator and encoder in most GANs become more and more perfect but the discriminator becomes more and more vanilla. In other words, the optimal discriminator is vanilla and useless in most cases.

A natural question is can we share weights between discriminator and encoder? Few of researches implement it successfully. As far as we know, the only one successfully done it is Introspective Variational Autoencoders (IntroVAEs, \cite{Huang2018IntroVAE}). But IntroVAEs are such complex that we are hard to reproduce their results.

In this paper, we develop a extremely concise approach to help the discriminator in common GANs become a encoder. More concretely, we develop a extra objective function for common GANs, and with this extra term the discriminator can not only distinguish whether the input image is real or not but also have ability to extract good features for the input image. We further discuss the degrees of freedom of discriminator, which reveals the principle of how this extra objective function works.

\section{Basic Analysis}

Most GANs are an alternating minimization w.r.t. generator $G$ and discriminator $D$:
\begin{equation}\begin{aligned}D =& \mathop{\arg\min}_{D} \mathbb{E}_{x\sim p(x), z\sim q(z)}\Big[f(D(x)) + g(D(G(z)))\Big]\\
G =& \mathop{\arg\min}_{G} \mathbb{E}_{z\sim q(z)}\Big[h(D(G(z)))\Big]
\end{aligned}\end{equation}
where $x\in \mathbb{R}^{n_x},z\in \mathbb{R}^{n_z}$, and $p(x)$ is the evidence distribution of real images and $q(z)$ is the distribution of noise\footnote{In our paper, $q(z)$ is $n_z$-dimensional standard normal distribution.}. $G: \mathbb{R}^{n_z} \to \mathbb{R}^{n_x}$ and $D: \mathbb{R}^{n_x} \to \mathbb{R}$ are generator and discriminator correspondingly. $f,g,h$ are some specific $\mathbb{R}\to\mathbb{R}$ functions. Sometimes we may need to add some normalization tricks or regularization terms into discriminator, such as \citep{Gulrajani2017Improved}, \citep{Roth2017Stabilizing}, \citep{Miyato2018Spectral} and \citep{Mescheder2018Which}.

An ordinary example is called vanilla GAN:
\begin{equation}\begin{aligned}D =& \mathop{\arg\min}_{D} \mathbb{E}_{x\sim p(x), z\sim q(z)}\Big[\text{sp}(-D(x)) + \text{sp}(D(G(z)))\Big]\\
G =& \mathop{\arg\min}_{G} \mathbb{E}_{z\sim q(z)}\Big[\text{sp}(-D(G(z)))\Big]
\end{aligned}\end{equation}
where $\text{sp}(x)=\log(1+e^x)$ is softplus function. It is not difficult to prove that the optimal discriminator of vanilla GAN is
\begin{equation}D^*(x)=\log \frac{p(x)}{q(x)}\end{equation}
where $q(x)=\int \delta(x - G(z))q(z)dz$ is the distribution of generated images. If it achieves ideal state $q(x)=p(x)$ we have $D^*(x)\equiv 0$.

Other GANs can be analyzed similarly and have similar conclusions. Therefore, we can see that the optimal discriminator of most GANs will degenerate into a trivial state (e.g. a constant), which means it is useless for us. This is a huge waste of the fitting ability of discriminator.

\section{Our Models}

First of all, we difine three operators for a vector:
\begin{equation}\text{avg}(z)=\frac{1}{n_z}\sum_{i=1}^{n_z} z_i,\quad \text{std}(z)=\sqrt{\frac{1}{n_z}\sum_{i=1}^{n_z} (z_i-\text{avg}(z))^2}, \quad \mathcal{N}(z)=\frac{z - \text{avg}(z)}{\text{std}(z)}\end{equation}
For all $n_z \geq 3$, $\left[\text{avg}(z), \text{std}(z), \mathcal{N}(z)\right]$ is functionally independent decomposition of vector $z$. We treat it as orthogonal decomposition analogically.

Then we decompose $D(x)$ as
\begin{equation}D(x)\triangleq T(E(x))\end{equation}
where $E$ is a $\mathbb{R}^{n_x} \to \mathbb{R}^{n_z}$ function and $T$ is a $\mathbb{R}^{n_z} \to \mathbb{R}$ function. We want to make $E(x)$ become a practical encoder for $x$. We found that it just need an additional reconstruction loss $- \rho(z, E(G(z)))$ on both generator and discriminator:
\begin{equation}\begin{aligned}T,E =& \mathop{\arg\min}_{T,E} \mathbb{E}_{x\sim p(x), z\sim q(z)}\Big[f(T(E(x))) + g(T(E(G(z)))) - \lambda \rho(z, E(G(z)))\Big]\\
G =& \mathop{\arg\min}_{G} \mathbb{E}_{z\sim q(z)}\Big[h(T(E(G(z))))  - \lambda \rho(z, E(G(z)))\Big]
\end{aligned}\label{eq:simplest-0}\end{equation}
where
\begin{equation}\rho(z, \hat{z})=\frac{\sum\limits_{i=1}^{n_z} (z_i - \text{avg}(z))(\hat{z}_i - \text{avg}(\hat{z}))/n_z}{\text{std}(z)\times \text{std}(\hat{z})}=\cos(\mathcal{N}(z), \mathcal{N}(E(G(z))))\end{equation}
is Pearson correlation between vector $z$ and $\hat{z}$.

Obiviously this auxiliary term can be embedded into any kind of GANs. Here we propose a simplest solution: use $\text{avg}(E(x))$ as discriminator directly:
\begin{equation}\begin{aligned}E =& \mathop{\arg\min}_{E} \mathbb{E}_{x\sim p(x), z\sim q(z)}\Big[f(\text{avg}(E(x))) + g(\text{avg}(E(G(z)))) - \lambda \rho(z, E(G(z)))\Big]\\
G =& \mathop{\arg\min}_{G} \mathbb{E}_{z\sim q(z)}\Big[h(\text{avg}(E(G(z)))) - \lambda \rho(z, E(G(z)))\Big]
\end{aligned}\label{eq:simplest}\end{equation}
Via this way, we can omit the network $T(\cdot)$ and leave no redundancy at all. Both $\eqref{eq:simplest-0}$ and $\eqref{eq:simplest}$ are called Orthogonal Generative Adversarial Networks (O-GANs) because they are based the orthogonal decomposition of discriminator.

\section{Arguments}

Some intuition might help understand why such a simple modification might work. The first thought of reconstruction objective might be $\Vert z - E(G(z))\Vert^2$ rather than $\rho(z, E(G(z)))$. However, $\Vert z - E(G(z))\Vert^2$ does not work and even destroy the original training process of GANs.

As defined above, $E(x)$ outputs one $n_z$-dim vector and $T(E(x))$ just outputs one scalar. In other words, as a discriminator, $T(E(x))$ must occupy one degree of freedom of $E(x)$ \footnote{and only need to occupy one degree of freedom in theory.}. If we add minimize $\Vert z - E(G(z))\Vert^2$, it will occupy all degrees of freedom of $E(G(z))$ \footnote{Because $\Vert z - E(G(z))\Vert^2=0$ if and only if $z = E(G(z))$.} and leave no freedom for discriminator. Conversely, $\rho(z, E(G(z)))$ is rrelevant with $\text{avg}(E(G(z)))$ and $\text{std}(E(G(z)))$, which means it leaves two degree of freedom of $E(G(z))$ for discriminator while minimizing $-\rho(z, E(G(z)))$. This is also why can we use $\text{avg}(E(x))$ as discriminator in $\eqref{eq:simplest}$.

Another argument is that we train the generator adversarially with $z \sim \mathcal{N}(0,I_{n_z})$. For a well-trained generator $G$, $G(z)$ can produce a high-quality image if and only if $z \sim \mathcal{N}(0,I_{n_z})$ (in theory). And if $z \sim \mathcal{N}(0,I_{n_z})$ then we have $\text{avg}(z)\approx 0$ and $\text{std}(z)\approx 1$. That is to say, for a well-trained generator $G$, if $G(z)$ is a high-quality image then the necessary condition is $\text{avg}(z)\approx 0$ and $\text{std}(z)\approx 1$. Thus, if $G(E(x))$ is a high-quality reconstruction of $x$, the necessary condition is $\text{avg}(E(x))\approx 0$ and $\text{std}(E(x))\approx 1$.

Consequently, we can say that the avg and std of $E(x)$ have been known for us, which means we do not need to fit $\text{avg}(E(x))$ and $\text{std}(E(x))$ by reconstruction loss \footnote{We have known they must be 0 and 1 correspondingly.}. So we can remove the avg and std from reconstruction loss -- actually, Pearson correlation can indeed be regarded as a mse loss removing avg and std:
\begin{equation}-\rho(z, E(G(z)))\sim \left\Vert \mathcal{N}(z) - \mathcal{N}(E(G(z)))\right\Vert^2\end{equation}
then the reconstrcution $\hat{x}$ of input $x$ is
\begin{equation}\hat{x}=G(\mathcal{N}(E(x)))\end{equation}
rather than $G(E(x))$.

Furthermore, this additional correlation loss can prevent GANs from mode collapse in theory. Because $E(G(z))$ can recover most of the information about $z$ after adding this term, which means generator $G$ can not be a degenerate function. $\rho(z, E(G(z)))$ can also be regraded as a lower bound of mutual information $I(Z, G(Z))$ between $Z$ and $G(Z)$ \citep{Xi2016InfoGAN}, which is to say that to minimize $-\rho(z, E(G(z)))$ is to maximize $I(Z, G(Z))$ and prevent mode collapse \footnote{According to \citep{Kumar2019Maximum}, to maximize $I(Z, G(Z))$ equals to maximize the entropy $H[G(Z)]$ of $G(Z)$. And maximizing $H[G(Z)]$ will increase diversity of $G(Z)$.}.

\section{Experiments}

\textbf{Datasets}\,\, We consider four datasets, namely CelebA HQ \citep{Karras2017Progressive}, FFHQ \citep{Karras2018A}, LSUN-bedroom \citep{yu15lsun} and LSUN-churchoutdoor \citep{yu15lsun}. We validate O-GANs on all datasets on $128\times 128$ resolution.

\textbf{Architecture}\,\, We use a DCGAN-based architecture \citep{Radford2015Unsupervised} for both generator $G$ and encoder $E$ (Table $\ref{tab:arc}$). The architecture are same no matter on which dataset. The kernel size of all Conv2D and Conv2DTranspose is $5\times 5$. And the negative slope coefficient of LeakyReLU is 0.2.

\linespread{1.25}
\begin{table}$$
\begin{array}{c|c}
\hline
\hline
\text{Encoder} & \text{Generator}\\
\hline

\begin{array}{cc}
\text{Layer} & \text{output\_shape}\\
\hline
\text{input\_image} & (128,128,3)\\
\text{Conv2D} &  (64, 64, 64) \\
\text{LeakyReLU} & (64, 64, 64) \\
\text{Conv2D} &  (32, 32, 128) \\
\text{BatchNormalization} &  (32, 32, 128) \\
\text{LeakyReLU} & (32, 32, 128) \\
\text{Conv2D} &  (16, 16, 256) \\
\text{BatchNormalization} &  (16, 16, 256) \\
\text{LeakyReLU} & (16, 16, 256) \\
\text{Conv2D} &  (8, 8, 512) \\
\text{BatchNormalization} &  (8, 8, 512) \\
\text{LeakyReLU} & (8, 8, 512) \\
\text{Conv2D} &  (4, 4, 1024) \\
\text{BatchNormalization} &  (4, 4, 1024) \\
\text{LeakyReLU} & (4, 4, 1024) \\
\text{Flatten} & 16384\\
\text{Dense} & 128\\
\\
\\
\end{array} & 

\begin{array}{cc}
\text{Layer} & \text{output\_shape}\\
\hline
\text{input\_noise} & 128\\
\text{Dense} &  16384 \\
\text{Reshape} & (4, 4, 1024) \\
\text{BatchNormalization} &  (4, 4, 1024) \\
\text{ReLU} & (4, 4, 1024) \\
\text{Conv2DTranspose} & (8, 8, 512) \\
\text{BatchNormalization} &  (8, 8, 512) \\
\text{ReLU} & (8, 8, 512) \\
\text{Conv2DTranspose} & (16, 16, 256) \\
\text{BatchNormalization} &  (16, 16, 256) \\
\text{ReLU} & (16, 16, 256) \\
\text{Conv2DTranspose} & (32, 32, 128) \\
\text{BatchNormalization} &  (32, 32, 128) \\
\text{ReLU} & (32, 32, 128) \\
\text{Conv2DTranspose} & (64, 64, 64) \\
\text{BatchNormalization} &  (64, 64, 64) \\
\text{ReLU} & (64, 64, 64) \\
\text{Conv2DTranspose} & (128, 128, 3) \\
\text{Tanh} & (128, 128, 3) \\
\end{array}

\\
\hline
\hline
\end{array}
$$
\caption{Architecture of Our DCGAN-base generator and encoder.}
\label{tab:arc}\end{table}

\textbf{Implementations}\,\, We implement $\eqref{eq:simplest}$ with $f(t)\equiv h(t)\equiv t$ and $g(t)\equiv-t$. An addtional differential formed regularization term \citep{Xiang2018Improving, su2018gan} is added on encoder. The final complete objective we use is as following
\begin{equation}\begin{aligned}E =& \mathop{\arg\min}_{E} \mathbb{E}_{x\sim p(x), z\sim q(z)}\Big[\text{avg}(E(x)) - \text{avg}(E(G(z))) + \lambda_1 R_{x,z} - \lambda_2 \rho(z, E(G(z)))\Big]\\
G =& \mathop{\arg\min}_{G} \mathbb{E}_{z\sim q(z)}\Big[\text{avg}(E(G(z))) - \lambda_2 \rho(z, E(G(z)))\Big]
\end{aligned}\label{eq:simplest-2}\end{equation}
where
\begin{equation}R_{x,z} = \frac{[\text{avg}(E(x)) - \text{avg}(E(G(z)))]^2}{\Vert x - G(z)\Vert^2}\end{equation}
and $\lambda_1 = 0.25/n_x, \lambda_2 = 0.5$. For faster speed, we do not use gradient penalty. The code is implemented in Keras \citep{chollet2015keras}, and available in my repository\footnote{\url{https://github.com/bojone/o-gan}}. We use the RMSprop optimizer, with a constant learning rate of $10^{-4}$ and $\beta = 0.99$ in both $G$ and $E$. We train $\eqref{eq:simplest-2}$ with one $E$ steps per $G$ step. It needs to run about 30k-50k iterations (about 12 hours on a GTX1060) to achieve the best result for each dataset.

\begin{figure}[H]
\centering
\subfigure[Random Samples on CelebA HQ]{
\includegraphics[width=6.75cm]{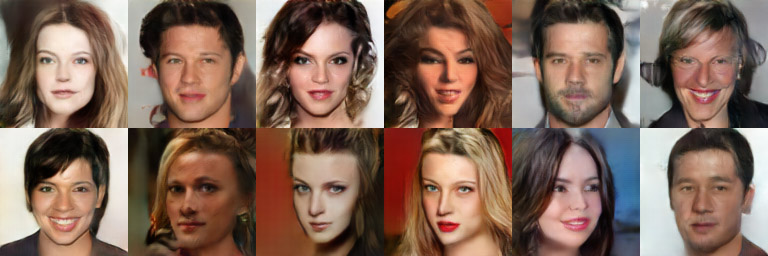}}\hspace{0.2cm}
\subfigure[Random Reconstructions on FFHQ]{
\includegraphics[width=6.75cm]{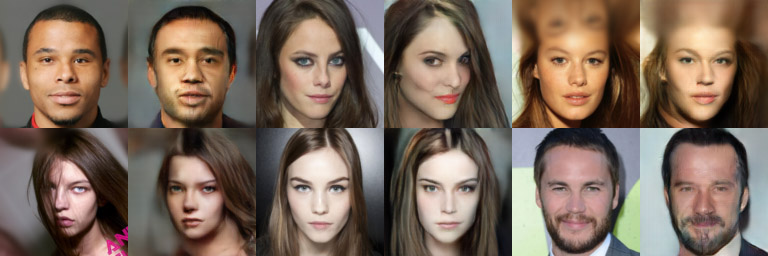}}

\subfigure[Linear Interpolations on CelebA HQ]{
\includegraphics[width=11cm]{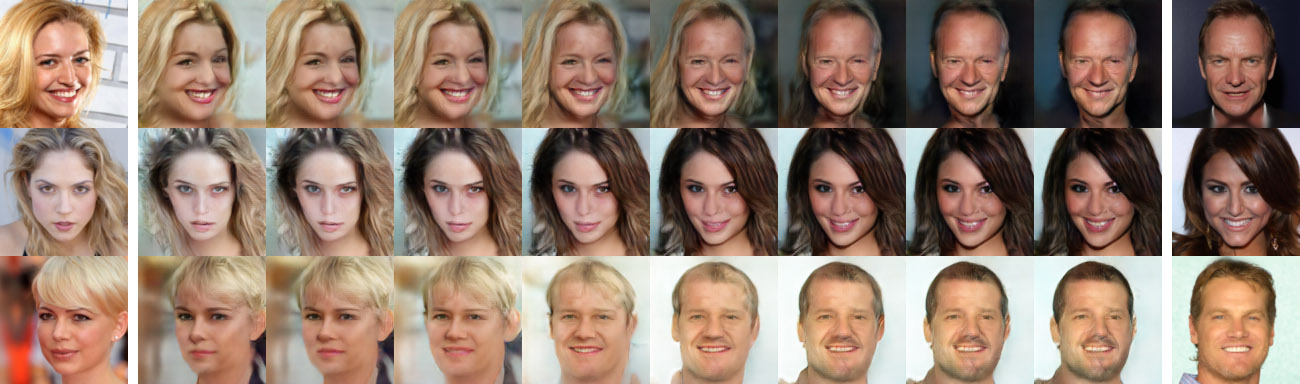}}
\caption{Results on CelebA HQ. Figure (a) is random generated images of $G(z)$; In figure (b), images at odd position is original images from training set and even position is its reconstruction correspondingly; In figure (c), left-most and right-most are original images, left-sencond and right-second are their reconstructions via our model, and the rest are linear interpolations of them in encoded space.}
\label{fig:celeba}
\end{figure}

\begin{figure}[H]
\centering
\subfigure[Random Samples on FFHQ]{
\includegraphics[width=6.75cm]{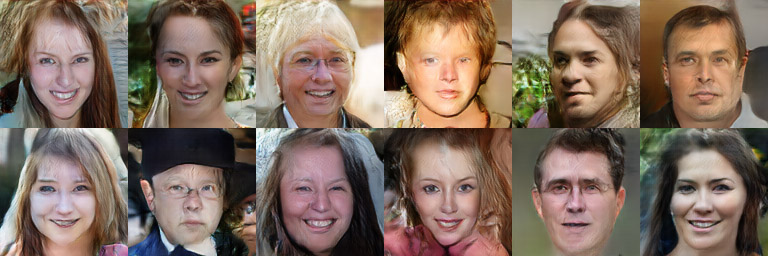}}\hspace{0.2cm}
\subfigure[Random Reconstructions on FFHQ]{
\includegraphics[width=6.75cm]{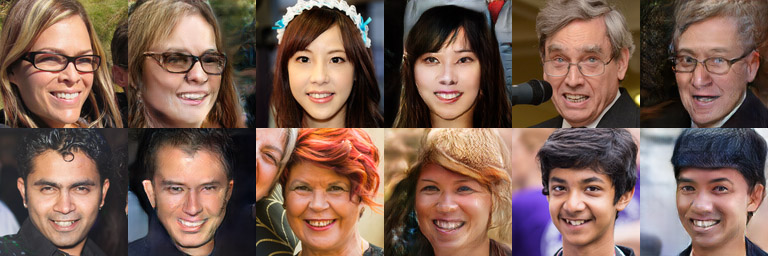}}

\subfigure[Linear Interpolations on FFHQ]{
\includegraphics[width=11cm]{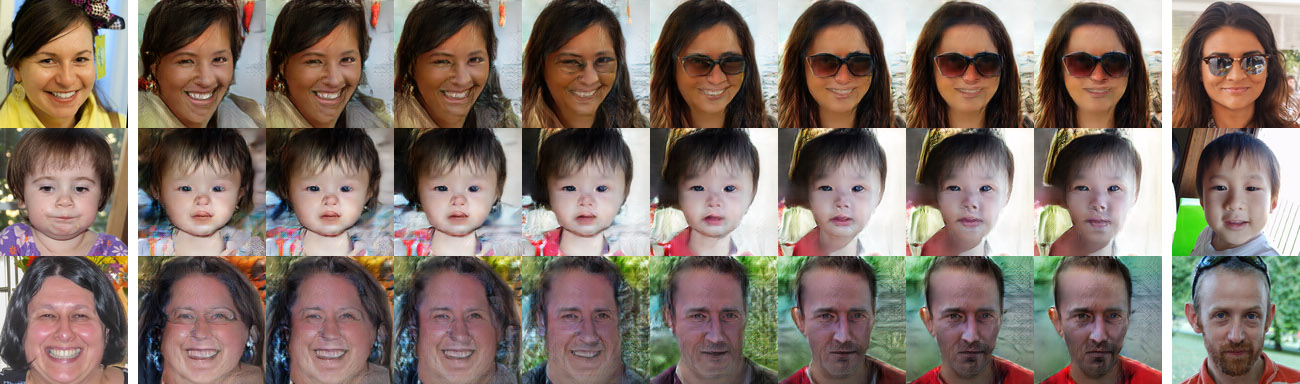}}
\caption{Results on FFHQ. Figure (a) is random generated images of $G(z)$; In figure (b), images at odd position is original images from training set and even position is its reconstruction correspondingly; In figure (c), left-most and right-most are original images, left-sencond and right-second are their reconstructions via our model, and the rest are linear interpolations of them in encoded space.}
\label{fig:ffhq}
\end{figure}

\begin{figure}[H]
\centering
\subfigure[Random Samples on LSUN-church]{
\includegraphics[width=6.75cm]{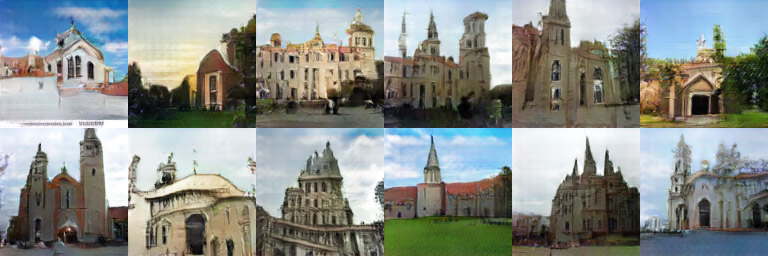}}\hspace{0.2cm}
\subfigure[Random Reconstructions on LSUN-church]{
\includegraphics[width=6.75cm]{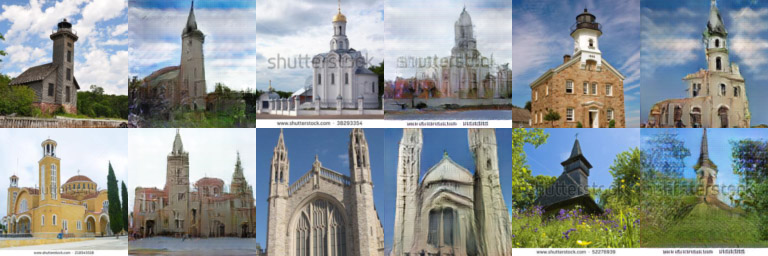}}

\subfigure[Linear Interpolations on LSUN-church]{
\includegraphics[width=11cm]{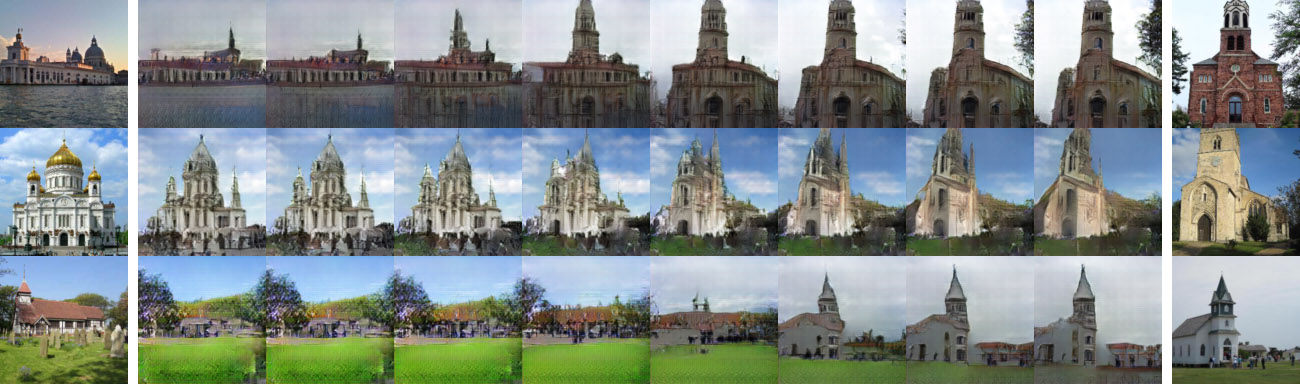}}
\caption{Results on LSUN-church. Figure (a) is random generated images of $G(z)$; In figure (b), images at odd position is original images from training set and even position is its reconstruction correspondingly; In figure (c), left-most and right-most are original images, left-sencond and right-second are their reconstructions via our model, and the rest are linear interpolations of them in encoded space.}
\label{fig:outdoor}
\end{figure}

\begin{figure}[H]
\centering
\subfigure[Random Samples on LSUN-bedroom]{
\includegraphics[width=6.75cm]{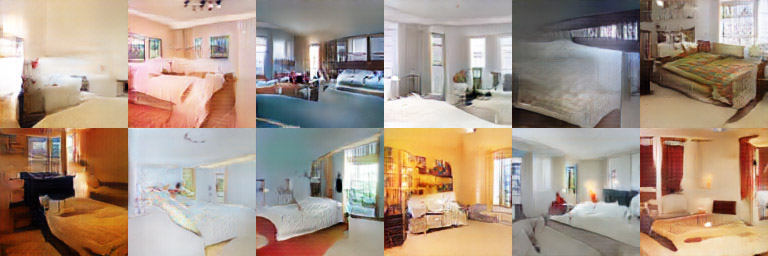}}\hspace{0.2cm}
\subfigure[Random Reconstructions on LSUN-bedroom]{
\includegraphics[width=6.75cm]{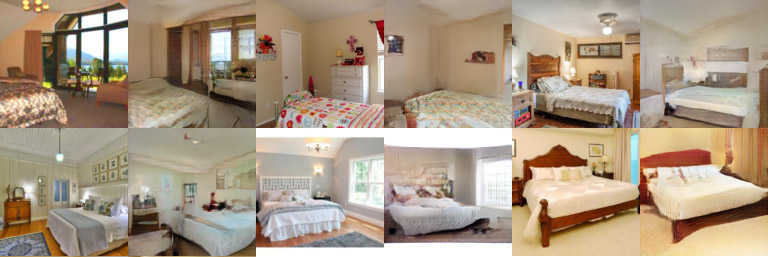}}

\subfigure[Linear Interpolations on LSUN-bedroom]{
\includegraphics[width=11cm]{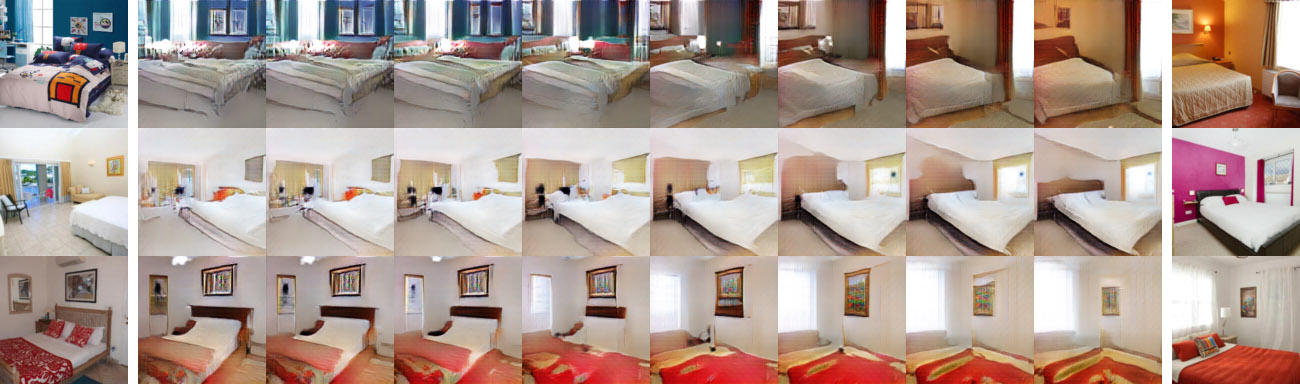}}
\caption{Results on LSUN-bedroom. Figure (a) is random generated images of $G(z)$; In figure (b), images at odd position is original images from training set and even position is its reconstruction correspondingly; In figure (c), left-most and right-most are original images, left-sencond and right-second are their reconstructions via our model, and the rest are linear interpolations of them in encoded space.}
\label{fig:bedroom}
\end{figure}

\textbf{Results}\,\, In theory, our extra objective can not improve the generated quality of original GANs. It just provides an approach to make original discriminator become an effective encoder. So the main results we care about is the reconstruction quality and disentangled learning ability of encoder. From subfigure (b) of figure $\ref{fig:celeba},\ref{fig:ffhq},\ref{fig:outdoor},\ref{fig:bedroom}$, we can see that the encoder actually keeps the core information of input image because of good reconstruction. From subfigure (c) we can see that the encoder obtains linearly separable features of input image because of natural transition of interpolation. Therefore, from all (a),(b) and (c), we can say that the additional correlation objective can actually provide a good encoder for GANs without loss in random generated quality. 

\section{Conclusions}

We have introduced Orthogonal Generative Adversarial Networks (O-GANs), a novel and extremely concise approach for training a generative adversarial network with auto-encoding ability. The new objective is just a simple modification of original GANs. As it makes the full use of all degrees of freedom of discriminator, we can obtain a nice encoder without increasing any training parameters and computational cost.

\newpage

\vspace{1.5cm}

\newpage

\vspace{1.5cm}

\newpage
\linespread{0.8}
\bibliographystyle{apacite}
\bibliography{o-gan.bib}

\end{document}